\documentclass[journal]{IEEEtran}
\IEEEoverridecommandlockouts

\usepackage{cite}
\usepackage{amsmath,amssymb,amsfonts}
\usepackage[ruled,linesnumbered]{algorithm2e}
\usepackage{graphicx}
\usepackage{textcomp}
\usepackage{xcolor}
\newtheorem{lemma}{\indent Lemma}
\def\BibTeX{{\rm B\kern-.05em{\sc i\kern-.025em b}\kern-.08em
    T\kern-.1667em\lower.7ex\hbox{E}\kern-.125emX}}
\usepackage[letterpaper, right=0.64in, left=0.63in, top=0.7in, bottom=1.04in]{geometry}
\begin{document}

\title{Learning-Based User Association for MmWave Vehicular Networks with Kernelized Contextual Bandits}

\author{{Xiaoyang He, Xiaoxia Huang\\ \textit{School of Electronics and Communication Engineering, Sun Yat-sen University} \\Email: hexy35@mail2.sysu.edu.cn, huangxiaoxia@mail.sysu.edu.cn }
\thanks{The work of Huang was supported by the National Natural Science Foundation of China under grant U22A2003 and 62271515.}
}

\maketitle
\pagestyle{empty}
\thispagestyle{empty}

\begin{abstract}
Vehicles require timely channel conditions to determine the base station (BS) to communicate with, but it is costly to estimate the fast-fading mmWave channels frequently.
Without additional channel estimations, the proposed Distributed Kernelized Upper Confidence Bound (DK-UCB) algorithm estimates the current instantaneous transmission rates utilizing past contexts, such as the vehicle's location and velocity, along with past instantaneous transmission rates.
To capture the nonlinear mapping from a context to the instantaneous transmission rate, DK-UCB maps a context into the reproducing kernel Hilbert space (RKHS) where a linear mapping becomes observable.
To improve estimation accuracy, we propose a novel kernel function in RKHS which incorporates the propagation characteristics of the mmWave signals.
Moreover, DK-UCB encourages a vehicle to share necessary information when it has conducted significant explorations, which speeds up the learning process while maintaining affordable communication costs.
\end{abstract}


\section{Introduction}

To support high data rates, low latency, and massive access, mmWave communication has emerged as a promising technology in vehicular communication networks \cite{9779354}.
Establishing connections between vehicles and BSs, known as user association, is challenging in mmWave vehicular networks.
MmWave signals struggle with obstacles and fast varying channel conditions introduced by vehicle mobility \cite{7390852}.
Additionally, the varying interference caused by concurrent transmissions from numerous vehicles exacerbates the uncertainty of channels.

The multi-armed bandits (MAB) framework is a promising model to capture dynamic channel conditions with affordable computation and communication costs.
Vehicles rely solely on past information to estimate instantaneous transmission rates of the uplink to each BS, without taking extra channel measurements at each period.
In the MAB framework, BSs and vehicles can be considered as arms and agents, respectively.
At each period, a learning agent takes actions (communicates with a BS), and interacts with the environment to receive a reward as the instantaneous transmission rate.
The contextual MAB (CMAB) framework is a powerful method to capture the non-stationary reward distributions in mmWave vehicular networks.
Sim et al. \cite{8472783} and Li et al. \cite{li2020smart} have applied the CMAB framework to model the non-stationary reward distributions caused by time-varying traffic congestion and blockages.
They partition the context space into small-scale hypercubes and explore all hypercubes, but it is ineffective given the large coverage area.
Although some work leverages the correlation between different hypercubes to eliminate unnecessary explorations and achieves fast convergence \cite{CC-UCB}, the number of hypercubes which need to be explored remains substantial.
Regression is another approach to estimate reward distributions for a CMAB framework, such as the LinUCB algorithm \cite{chu2011contextual}.
However, LinUCB assumes that the reward follows Lipschitz-continuity and exhibits a linear relationship with the context, which is too restrictive for mmWave vehicular channels.

In this paper, we propose a distributed kernelized upper confidence bound (DK-UCB) algorithm for user association in mmWave vehicular networks.
DK-UCB can estimate the reward given a context by learning the mapping from a context to the corresponding reward.
Recognizing that the mapping is nonlinear in mmWave vehicular networks, DK-UCB uses kernel methods to map the context into the Hilbert space where a linear mapping becomes observable.
In kernel methods, the kernel function, which evaluates the similarity between two contexts, plays an important role in estimation accuracy.
We devise a novel kernel function to capture the unique propagation characteristics of mmWave vehicular networks, such as path loss, Doppler spread, and interference.
Further, the proposed model can be formulated as a multi-agent MAB (MA-MAB) problem.
Sharing information among vehicles can accelerate the learning process, but frequent communication between vehicles and BSs is expensive in mmWave vehicular networks.
In DK-UCB, a vehicle only shares necessary information when it has conducted significant explorations, ruling out unnecessary communication.

The rest of this paper is organized as follows. 
Section II provides the system model.
The DK-UCB algorithm and simulation results are introduced in Section III and Section IV.
Section V finally concludes the paper.

\section{System Model}
To address user association in mmWave vehicular networks, we first introduce the mobility model of vehicles, along with the channel models.
Following this, we formulate the optimization problem based on the network model.

\subsection{Mobility Model and Channel Model}
Let $T \in \mathbb{N}$ denote a finite time horizon, vehicles update their locations at each period $t = 1,\dots,T$.
Let $\mathbb{U}(t) = \{1,\dots,i,\dots,|\mathbb{U}(t)|\}$ and $v_{t,i}$ represent the set of vehicles and the velocity of vehicle $i$ at $t$, respectively.
The arrival of vehicles at each period follows a Poisson distribution with density $\lambda$.
Let $\mathbb{B} = \{1,\dots,j\dots,N_{BS}\}$ indicate the set of BSs.
Each vehicle is assumed to be equipped with a rectangular antenna array according to the 3GPP TR 38.901 standard \cite{3gpp2018study}.
A massive antenna array is deployed on each BS, enabling the BS to serve multiple vehicles simultaneously.

Consider BS $j$ and vehicle $i$.
When BS $j$ serves vehicle $i$, the channel gain between vehicle $i$ and BS $j$ is
\begin{equation}
	\label{h_ij_0}
	h_{i,j}(t) = {(\mathbf{w}^r_{i,j})} ^*\mathbf{H}_{i,j}(t)\mathbf{w}^t_{i,j}.
\end{equation}
Here, $\mathbf{H}_{i,j}(t)$ is the channel matrix between vehicle $i$ and BS $j$.
$\mathbf{w}^r_{i,j}$ and $\mathbf{w}^t_{i,j}$ are the receiving and transmitting beamforming weights, respectively.

Considering another vehicle $k$ communicates to BS $l$ over an uplink.
At period $t$, the perceived interference to BS $j$ from vehicle $k$ while serving vehicle $i$ is
\begin{equation}
	\label{h_ij}
	\widetilde{h}^{k,l}_{i,j}(t) = {(\mathbf{w}^r_{i,j})} ^*\mathbf{H}_{k,j}(t)\mathbf{w}^t_{k,l},
\end{equation}
where $\mathbf{w}^t_{k,l}$ is the transmitting beamforming weight of vehicle $k$.
Assume each vehicle transmits at power $P_v$, the interference plus noise at BS $j$ while serving vehicle $i$ is
\begin{equation}
	\label{I_ij}
	I_{i,j}(t) = \Big|\sum_{k \in (\mathbb{U}(t) \backslash i)} \sum_{l \in \mathbb{B}}  P_v \widetilde{h}^{k,l}_{i,j}(t) \mathcal{I}_{k,l}(t) \Big| ^ 2 + N_oW,
\end{equation}
where $W$ is the bandwidth and $N_o$ is the thermal noise power density.
$\mathcal{I}_{k,l}(t)$ is the indicator function.
If vehicle $k$ communicates to BS $l$, $\mathcal{I}_{k,l}(t) = 1$.
Otherwise, $\mathcal{I}_{k,l}(t) = 0$.

\subsection{Optimization Problem}
According to (\ref{h_ij}) and (\ref{I_ij}), the instantaneous transmission rate of vehicle $i$ communicating to BS $j$ is
\begin{equation}
	\label{R_ij}
	R_{i,j}(t) = W\log_2\big(1 + P_v|h_{i,j}(t)|^2/I_{i,j}(t)\big).
\end{equation}

At each period $t$, one vehicle can only communicate with one BS.
Let $\beta_i(t)$ indicate the index of the serving BS for vehicle $i$.
The association between vehicles and BSs can be defined as $\boldsymbol{\beta}(t) \triangleq [\beta_1(t),\dots,\beta_i(t),\dots,\beta_{|\mathbb{U}(t)|}(t)].$
The ultimate objective of our study is to identify an association vector which maximizes the total instantaneous transmission rate of all vehicles.
According to (\ref{R_ij}), the optimization problem at period $t$ is formulated as 
\begin{equation}
	\label{formulate_semi_ACK_UCB}
	\begin{aligned}
		&\max_{\boldsymbol{\beta}} \quad \sum_{i \in \mathbb{U}(t)}R_{i,\beta_i(t)}(t) = \sum_{i \in \mathbb{U}(t)}  W\log_2(1 + \frac{P_v|h_{i,\beta_i(t)}(t)|^2}{I_{i,\beta_i(t)}(t)})\\
		&\begin{array}{r@{\quad}r@{}l@{\quad}l}
			s.t. &\sum_{j=1}^{N_{BS}}\mathcal{I}_{i,j}(t)& = 1, \forall i \in \mathbb{U}(t).\\
		\end{array}
	\end{aligned}
\end{equation}
The constraint in (\ref{formulate_semi_ACK_UCB}) indicates that a vehicle can only connect to one BS but a BS can serve multiple vehicles simultaneously.
The optimization problem formulated in (\ref{formulate_semi_ACK_UCB}) can be formulated as the Generalized Proportional Fairness (GPF1) problem.
The GPF1 problem can be reduced from the NP-hard 3-dimensional matching problem \cite{ramjee2006generalized}, so the problem in (\ref{formulate_semi_ACK_UCB}) is also NP-hard.

\section{Distributed Kernelized Upper Confidence Bound Algorithm}
\label{section DK-UCB}

In this section, we begin by providing an overview of the proposed DK-UCB algorithm.
Next, we describe the setting of the kernelized contextual bandits and the MA-MAB framework used in the DK-UCB algorithm in detail.
Finally, we derive an upper bound of the internal regret and communication cost.
\subsection{Framework of DK-UCB Algorithm}
\begin{figure}[htpb]
	\centering\includegraphics[scale=0.53]{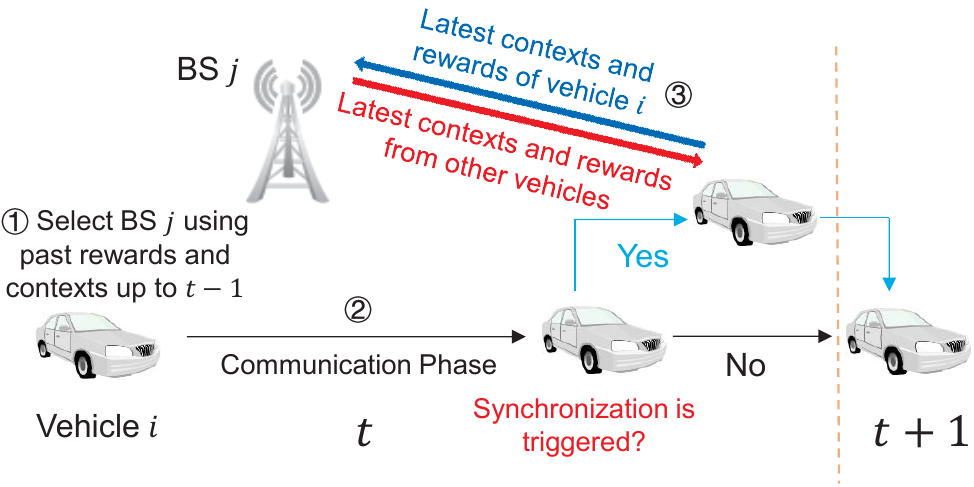}
	\caption{Architecture of the DK-UCB algorithm at period $t$.}
	\label{algorithm_structure}
\end{figure}
Consider a vehicle $i$, the DK-UCB algorithm operates as depicted in Fig.~\ref{algorithm_structure}.
At period $t$, vehicle $i$ requires estimated transmission rates for BSs to choose the one with a high transmission rate.
The transmission rates are non-stationary in mmWave vehicular networks but the MAB framework requires reward distributions to be stationary.
The non-stationary transmission rates are determined by many factors, such as the channel's geometrical characteristics, Doppler spread and interference according to (\ref{R_ij}).
These elements are affected by the vehicle's location, velocity, and the number of concurrent transmissions, which are non-stationary processes.
To capture the non-stationary characteristics, the CMAB framework is utilized in the DK-UCB algorithm.
The context in DK-UCB is defined as the vehicle's location, velocity, and the number of concurrent transmissions of BS $j$ for the last period of vehicle $i$ communicating with BS $j$.
Within each context, similar location and velocity lead to comparable channel geometrical characteristics and Doppler spread, while the comparable number of concurrent transmissions leads to comparable interference.
Therefore, the transmission rate can be considered drawn from a stationary reward distribution for a defined context.
Using kernel methods, vehicle $i$ can estimate rewards for each BS by learning the nonlinear mapping from a given context to the corresponding reward.

More available contexts and rewards help improve the accuracy of the estimation in kernel methods.
Information sharing can increase available contexts and rewards, but frequent communication can introduce significant overhead.
In DK-UCB, when a vehicle has conducted significant explorations, the sharing process, referred to as synchronization, is triggered.
If the synchronization is not triggered, the vehicle takes no further action, thereby eliminating unnecessary synchronizations.

\subsection{Kernelized Contextual Bandits}

At period $t$, each vehicle $i$ chooses a BS $a_{t,i}$ from a candidate BS set $\mathcal{A}_{t,i} \in \mathbb{B}$ to communicate with.
Recall that $\mathbb{B}$ is the set of BSs in the network.
$\mathcal{A}_{t,i}$ includes the BSs whose distance from the vehicle is below $R_{max}$, because the BSs located far from the vehicle exhibit poor channel conditions due to the severe path loss.
Then, vehicle $i$ receives a reward $y_{t,i} = f(\mathbf x_{t,i}(a_{t,i})) + \eta$.
$\mathbf x_{t,i}(a_{t,i}), f$ and $\eta$ denote the context, reward function and noise, respectively.
The reward is also determined by other factors, such as the locations of interfering vehicles, which introduce the noise.
Let $d$ indicate the dimension of the context.
The context $\mathbf x_{t,i}(a_{t,i})$ for BS $a = a_{t,i}$ is defined as $\mathbf x_{t,i}(a_{t,i}) = (a_{t,i},\theta_{t,i}(a), L_{t,i}(a), f_{t,i}(a), N_{t,i}(a)) \in \mathbb{R}^{d\times1}$, incorporating the factors determining the reward, which are described as follows.
\begin{itemize}
	\item $a_{t,i}$: Identity of a BS, i.e., an arm.
	\item $\theta_{t,i}(a)$: Orientation of the vector from arm $a$ to vehicle $i$ with respect to the positive x-axis.
	\item $L_{t,i}(a)$: Distance between vehicle $i$ and BS $a$.
	\item $f_{t,i}(a)$: Maximum Doppler spread.
	The mmWave propagation is dominated by the Line-of-Sight (LOS) propagation, so we define $f_{t,i}(a) = \frac{v_{t,i}(a)}{\lambda_f}$, where $\lambda_f$ is the wavelength and $v_{t,i}(a)$ is the component of $v_{t,i}$ along the tangent with respect to BS $a$.
	\item $N_{t,i}(a)$: Number of concurrent transmissions of BS $a$ for the last period of vehicle $i$ communicating with BS $a$.
\end{itemize}

DK-UCB uses current estimated mean reward $\hat{\mu}_{t,i}(\mathbf x_{t,i}(a))$ and standard deviation $\hat{\sigma}_{t,i}(\mathbf x_{t,i}(a))$ to construct a UCB index \cite{lai1985asymptotically}.
Then, the BS with the highest UCB index is selected.
\begin{equation}
	\label{arm_selection}
	a_{t,i} = {\arg\max}_{a \in \mathcal{A}_{t,i}} \hat{\mu}_{t,i}(\mathbf x_{t,i}(a)) + \alpha\hat{\sigma}_{t,i}(\mathbf x_{t,i}(a)).
\end{equation}
Parameter $\alpha$ determines whether the algorithm focuses more on exploration or exploitation.
Now it remains challenging to derive $\hat{\mu}_{t,i}(\mathbf x_{t,i}(a))$ and $\hat{\sigma}_{t,i}(\mathbf x_{t,i}(a))$.

It is essential to learn the relationship between the context associated with each action and the expected reward.
However, the reward function in (\ref{R_ij}) is nonlinear with respect to the context, so some famous algorithms such as LinUCB are not applicable. 
Kernel methods can potentially capture the nonlinear mapping from a context to the corresponding reward, relying solely on the similarity between contexts.
Given a mapping $\phi:\mathbb{R}^d \rightarrow \mathcal H$, which transforms the context into a (potentially infinite-dimensional) Hilbert space where a linear mapping from a context to the corresponding reward becomes observable.
Here $\mathbb{R}^d$ is the primal context space and $\mathcal H$ is the associated reproducing kernel Hilbert space (RKHS).
Since $\mathcal H$ may be infinite, finding the explicit form of the mapping $\phi$ is challenging.
Fortunately, Valko et al. have derived the estimated reward $\hat{\mu}_{t,i}(\mathbf x_{t,i}(a))$ and confidence interval $\hat{\sigma}_{t,i}(\mathbf x_{t,i}(a))$ using only the kernel function, without knowing $\phi$ \cite{valko2013finite}.
Consider two contexts $\mathbf x_{t,i}(a)$ and $\hat{\mathbf{x}}_{\hat t,j}(\hat a)$,
the kernel function evaluates the similarity between two contexts $\mathbf x_{t,i}(a)$ and $\hat{\mathbf{x}}_{\hat t,j}(\hat a)$, which is defined by $\kappa(\hat{\mathbf{x}}_{\hat t,j}(\hat a),\mathbf x_{t,i}(a)) := \phi(\hat{\mathbf{x}}_{\hat t,j}(\hat a))^\top\phi(\mathbf x_{t,i}(a)), \forall \hat{\mathbf{x}}_{\hat t,j}(\hat a),\mathbf x_{t,i}(a)\in \mathbb{R}^d$.

Let $\mathbf S_i(t) \in \mathbb{R} ^ {|\mathbf S_i(t)| \times d}$ and $\mathbf R_i(t) \in \mathbb{R} ^ {|\mathbf S_i(t)| \times 1}$ represent the set of the available contexts and rewards of vehicle $i$ up to period $t$.
The available contexts and rewards include both locally sampled data and received contexts and rewards from the BS.
Let $\lambda_k$ denote the regularization parameter, the estimated mean reward and standard deviation given a context $\mathbf x_{t,i}(a)$ are \cite{valko2013finite}
\begin{equation}
	\label{estimated_reward_confidence}
	\begin{aligned}
		\hat\mu_{t,i}(\mathbf x_{t,i}(a)) &= \mathbf{K}_{\mathbf x_{t,i}(a), \mathbf S_i(t)}^\top(\mathbf{K}_{\mathbf S_i(t),\mathbf S_i(t)} + \lambda_k\mathbf I)^{-1}\mathbf R_i(t)\\
		\hat\sigma_{t,i}(\mathbf x_{t,i}(a)) &= \lambda_k^{-1/2} \Big(\kappa(\mathbf x_{t,i}(a),\mathbf x_{t,i}(a)) - \mathbf{K}_{\mathbf x_{t,i}(a), \mathbf S_i(t)}^\top\times\\
		&(\mathbf{K}_{\mathbf S_i(t),\mathbf S_i(t)} + \lambda_k\mathbf I)^{-1}\mathbf{K}_{\mathbf x_{t,i}(a), \mathbf S_i(t)}\Big)^{\frac 12},
	\end{aligned}
\end{equation}
where,
\begin{equation*}
	\begin{aligned}
		&\mathbf{K}_{\mathbf x_{t,i}(a), \mathbf S_i(t)} = [\kappa\big(\mathbf x_{t,i}(a),\hat{\mathbf{x}}_{\hat t,j}(\hat a)\big)]^\top_{\hat{\mathbf{x}}_{\hat t,j}(\hat a)\in \mathbf S_i(t), \mathbf x_{t,i}(a)\notin \mathbf S_i(t)} \\
		&\mathbf{K}_{\mathbf S_i(t),\mathbf S_i(t)} = [\kappa\big(\bar{\mathbf x}_{\bar t,i}(\bar a),\hat{\mathbf{x}}_{\hat t,j}(\hat a)\big)]_{\bar{\mathbf x}_{\bar t,i}(\bar a), \hat{\mathbf{x}}_{\hat t,j}(\hat a) \in \mathbf S_i(t)}\\
		&\mathbf{K}_{\mathbf x_{t,i}(a), \mathbf S_i(t)} \in \mathbb{R} ^ {|\mathbf S_i(t)| \times 1}, \mathbf{K}_{\mathbf S_i(t),\mathbf S_i(t)} \in \mathbb{R} ^ {|\mathbf S_i(t)| \times |\mathbf S_i(t)|}
	\end{aligned}.
\end{equation*}

Now investigate the kernel function.
The transmission rate is determined by blockage, path loss, Doppler spread, and interference.
Therefore, evaluating the similarity of each element individually is more accurate rather than relying on a single kernel function. 
If $a = \hat a$, the kernel function is defined as,
\begin{equation}
	\label{kernel_proposed}
	\begin{aligned}
		\kappa\big(\mathbf x_{t,i}(a),\hat{\mathbf{x}}_{\hat t,j}(\hat a)\big) = & k_{\theta}(\mathbf \theta_{t,i}(a),\hat{\mathbf{\theta}}_{\hat t,j}(\hat a))k_L(L_{t,i}(a),\hat{L}_{\hat t,j}(\hat a)) \times\\&k_f(f_{t,i}(a),\hat{f}_{\hat t,j}(\hat a))k_N(N_{t,i}(a),\hat{N}_{\hat t,j}(\hat a)).
	\end{aligned}
\end{equation}
Otherwise, the kernel function is 0 because the reward distributions for different BSs are not correlated.
The four components of the devised kernel function correspond to the similarity of the two contexts in blockage, path loss, Doppler spread, and interference.
In the following, we show that our proposed kernel function in (\ref{kernel_proposed}) is valid, satisfying Mercer's Theorem \cite{mercer1909xvi}.
In the RKHS theory, Mercer's Theorem states that a kernel function $\kappa(\cdot,\cdot)$ is a valid kernel if $\kappa(\cdot,\cdot)$ is symmetric and its corresponding kernel matrix is positive semi-definite.
\subsubsection{Blockage and Path Loss}
Denote $\Delta \theta$ as $|\mathbf \theta_{t,i}(a) - \hat{\mathbf{\theta}}_{\hat t,j}(\hat a)|$.
The blockage probability is proportional to $\cos\Delta \theta$ given a small $\Delta \theta$ \cite{8114332}.
Thus, the similarity of blockage conditions is defined as
\begin{equation*}
	k_{\theta}(\mathbf \theta_{t,i}(a),\hat{\mathbf{\theta}}_{\hat t,j}(\hat a)) = 
	\begin{cases}
		\cos\Delta \theta & \Delta \theta < \frac\pi2\\
		0 & \text{otherwise}
	\end{cases}.
\end{equation*}
A small $\Delta \theta$ leads to a high probability that the two contexts experience similar blockage conditions.
Given a large $\Delta \theta > \frac\pi2$, we assume that there is no similarity regarding blockage conditions.
Let $\psi(x) = [\cos x, \sin x]$ and $\langle \cdot, \cdot \rangle$ denote inner product, $\cos(\Delta \theta) = \langle \psi(\mathbf \theta_{t,i}(a)), \psi(\hat{\mathbf{\theta}}_{\hat t,j}(\hat a)) \rangle$.
The linear kernel $\langle \cdot, \cdot \rangle$ is a valid kernel function, so $k_{\theta}(\cdot,\cdot)$ is also valid. 

The distance to the BS also determines the blockage probability.
Denote $\Delta L$ as $|L_{t,i}(a)-\hat{L}_{\hat t,j}(\hat a)|$.
If $\Delta L$ is large, the two contexts may not experience similar blockage conditions even with a small $\Delta \theta$.
Moreover, the discrepancy of the path loss between the two contexts is dominated by $\Delta L$.
We deploy the Gaussian kernel to evaluate the similarity for blockage and path loss with respect to distance, because Gaussian kernels are effective at capturing nonlinear characteristics \cite{shawe2004kernel},
\begin{equation*}
	k_L(L_{t,i}(a),\hat{L}_{\hat t,j}(\hat a)) = \exp\big(-{\Delta L}^2/(2\sigma^2_{L})\big).
\end{equation*}
Here, $\sigma^2_{L}$ is the parameter of the Gaussian kernel.

\subsubsection{Doppler Spread}
We deploy the exponential kernel to capture the similarity over Doppler spread, which is less sensitive to variables than the Gaussian kernel \cite{paclik2000road}.
Let $\sigma_f$ denote the parameter of the exponential kernel,
\begin{equation*}
	k_f(f_{t,i}(a),\hat{f}_{\hat t,j}(\hat a)) = \exp(-|f_{t,i}(a) - \hat{f}_{\hat t,j}(\hat a)|/\sigma_f).
\end{equation*}

\subsubsection{Interference}
The interference is dominated by the number of concurrent transmissions.
Consider $N$ vehicles communicating with a BS.
If a new vehicle joins in, we assume that the transmission rate of $N$ vehicles to the same BS decreases by $\frac1{\sigma_N}$.
Therefore, we capture the similarity of two contexts in interference by deploying the triangular kernel \cite{genton2001classes}
\begin{equation*}
	k_N(N_{t,i}(a),\hat{N}_{\hat t,j}(\hat a)) = (1-|N_{t,i}(a) - \hat{N}_{\hat t,j}(\hat a)|/\sigma_N)^+,
\end{equation*}
where $(x)^+ = \max(x,0)$.

The following lemma shows our proposed kernel function in (\ref{kernel_proposed}) is a valid kernel function.

\begin{lemma}
	\label{kernel_rule}
	If $\kappa_1$ and $\kappa_2$ are valid kernel functions, $\kappa\big(\mathbf x_{t,i}(a),\hat{\mathbf{x}}_{\hat t,j}(\hat a)\big) = \kappa_1\big(\mathbf x_{t,i}(a),\hat{\mathbf{x}}_{\hat t,j}(\hat a)\big)\kappa_2\big(\mathbf x_{t,i}(a),\hat{\mathbf{x}}_{\hat t,j}(\hat a)\big)$ is also valid kernel function \cite{shawe2004kernel}.
	Therefore, the product of multiple valid kernel functions remains a valid kernel function.
\end{lemma}

\subsection{MA-MAB framework}
A straightforward information update for bandit learning is through immediate sharing, where each agent shares every new context and reward with others at once.
Nevertheless, immediate sharing introduces an update storm in mmWave vehicular networks.
To reduce the communication overhead while improving estimation accuracy, a vehicle triggers a synchronization when considerable explorations are conducted.

Let $\mathbf S_i^a(t)$ and $\mathbf R_i^a(t)$ indicate the available contexts and rewards of vehicle $i$ communicating with BS $a$ up to period $t$, and $t^{i,a}_{syn}$ denote the timestamp of the last synchronization between vehicle $i$ and BS $a$.
The trigger-event for vehicle $i$ and BS $a$ at period $t$ is defined as,
\begin{equation}
	\label{synchronization_event}
	\begin{aligned}
		&\mathcal{U}_{t,i}(D) = \big\{(|\mathbf S_i^a(t)| - |\mathbf S_i^a(t^{i,a}_{syn})|)\times\\&\log\Big(\frac{\det(\mathbf I + \lambda_k^{-1}\mathbf K_{\mathbf S_i^a(t),\mathbf S_i^a(t)})}{\det(\mathbf I + \lambda_k^{-1}\mathbf K_{\mathbf S_i^a(t)\backslash \mathbf S_i^a(t^{i,a}_{syn}),\mathbf S_i^a(t)\backslash \mathbf S_i^a(t^{i,a}_{syn})})}\Big) > D \big\}.
	\end{aligned}
\end{equation}
Here $D$ is a coefficient to achieve a trade-off between communication cost and estimation accuracy.
The event $\mathcal{U}_{t,i}(D)$ is true if vehicle $i$ has accumulated plenty of new rewards and contexts with low similarity to the past contexts for BS $j$ since $t^{i,a}_{syn}$.

During the synchronization, vehicle $i$ uploads the new contexts and rewards to BS $a$.
Although vehicle $i$ can receive all the latest contexts and rewards from BS $a$, the received data scales with the number of vehicles, introducing significant communication costs given a large density of vehicles.
In DK-UCB, BS $a$ identifies the context subspace $\mathbb{S}(\mathbf x_{t,i}(a))$ that vehicle $i$ is likely to sample from in the short succeeding periods and sends only the contexts within this context space along with the rewards back to vehicle $i$.
The subspace is defined as
\begin{equation}
	\label{subspace_definition}
	\mathbb{S}(\mathbf x_{t,i}(a)) = \{\mathbf x \in \mathbb{R}^d : dis(\mathbf x,\mathbf x_{t,i}(a)) < R_p\},
\end{equation}
where $dis(\mathbf x,\mathbf x_{t,i}(a))$ indicates the distance between the locations of the two contexts and $R_p$ is a coefficient determining the size of the subspace.
Because the contexts outside and inside the subspace bear minimal similarity, the contexts outside the subspace contribute negligible information for estimation in the short succeeding periods.
Consider a succeeding period $\hat t$, if the sampled context resides outside the subspace $\mathbb{S}(\mathbf x_{t,i}(a))$, a synchronization is required for vehicle $i$ to receive the latest contexts within the latest subspace $\mathbb{S}(\mathbf x_{\hat t,i}(a))$.
The proposed DK-UCB algorithm is presented in Algorithm \ref{DK-UCB_ALG}.

\begin{algorithm}
	\caption{DK-UCB}
	\label{DK-UCB_ALG}
	\KwIn{Period $t$, coefficient $D$}
	\For{vehicle $i = 1,2,\dots,\mathbb{U}(t)$}{
		Obtain context $\mathbf{x}_{t,i}(j)$ for each BS $j\in\mathcal{A}_{t,i}$
		
		Communicate with $a = a_{t,i}$ calculated by (\ref{arm_selection}) and receive a reward
		
		\If{$\mathcal{U}_{t,i}(D)$ is true or $dis(\mathbf{x}_{t,i}(a),\mathbf x_{t^{i,a}_{syn},i}(a)) > R_p$}{
			Sends contexts $\mathbf S_i^a(t)\backslash \mathbf S_i^a(t^{i,a}_{syn})$ and rewards $\mathbf R_i^a(t)\backslash \mathbf R_i^a(t^{i,a}_{syn})$ to BS $a$
			
			Receives contexts $\{\cup_{k\neq i}^{\mathbb{U}(t)}\mathbf S_k^a(t^{k,a}_{syn})\} \cap \mathbb{S}(\mathbf x_{t,i}(a))$ along with the rewards from BS $a$
			
			Set $t^{i,a}_{syn} = t$ 
		} 
	}
\end{algorithm}


\subsection{Regret and Communication Cost Analysis}
An MA-MAB is a game involving two agents: agent $i$ and the set of other agents, whose joint action profile influences the reward received by agent $i$.
Each agent $i$ aims to minimize the internal regret \cite{stoltz2005internal}, which compares the expected reward of the current strategy with that of the best action at each period.
The upper bound of the internal regret for DK-UCB is given below. 
\begin{lemma}[\cite{wang2019distributed,li2022communication} ]
	\label{lemma_DK_UCB}
	With threshold $D = T/\gamma_{T}$ and the norm of the optimal reward vector $\|\theta_*\| \le \infty$, $\alpha = \sqrt{\lambda_k}\|\theta_*\| + R\sqrt{4\log T/\delta + 2\log\det(\mathbf I + \lambda_k^{-1}\mathbf K_{\mathbf S_i^j(t),\mathbf S_i^j(t)})}$, the cumulative internal regret of the DK-UCB algorithm for a vehicle is upper bounded by $\mathcal{O}(\sqrt{T}(\|\theta_*\|\sqrt{\gamma_{T}}+ \gamma_{T}))$ with probability at least $1 - \delta$.
\end{lemma}

Here, $\gamma_{T}$ denotes the maximum information gain, which is dependent on the kernel function and defined as  $\gamma_{T}:=\max_{\mathbf{S} \subset \mathbb{R}^d:|\mathbf{S}| = T}\log(\det(\mathbf{K}_{\mathbf{S}, \mathbf{S}}/\lambda_k + \mathbf I))/2$.
According to Lemma \ref{lemma_DK_UCB}, the regret of each agent can be bounded by $\mathcal{O}(\sqrt{T}(\|\theta_*\|\sqrt{\gamma_{T}}+ \gamma_{T}))$.



Let $\bar{U}$ represent the expected number of vehicles within the network.
Up to period $T$, each vehicle uploads and downloads data at most $\mathcal{O}(Td)$ and $\mathcal{O}(\bar{U}Td)$, so the communication cost is upper bounded by  $\mathcal{O}(\bar{U}^2Td)$.

\section{Numerical Results}

The roads and obstructions in the simulation scenario are accurately modeled based on the layout of Yuexiu District, Guangzhou, China.
The simulation area is defined by the coordinates of the upper right and bottom left corners of this region, specifically (23.1406N, 113.2770E) and (23.1280N, 113.2613E), respectively.
Vehicle mobility introduces strong Doppler spreads and time-varying channel conditions, resulting in a short channel coherence time.
We employ the Clustered Delay Line (CDL) channel model with ray tracing  \cite{3gpp2018study} to simulate the rapidly changing communication channel based on landscape data from OpenStreetMap \cite{OpenStreetMap}, considering the static blockages caused by buildings.
We assume for the worst case that all vehicles share the same bandwidth, leading to both intra-cell and inter-cell interference.
The velocity $v_{t,i}$ follows a uniform distribution on [20km/h, 80km/h] and the carrier frequency of the mmWave signal is 28GHz.
%

\begin{figure}[hpb]
	\centering\includegraphics[scale=0.55]{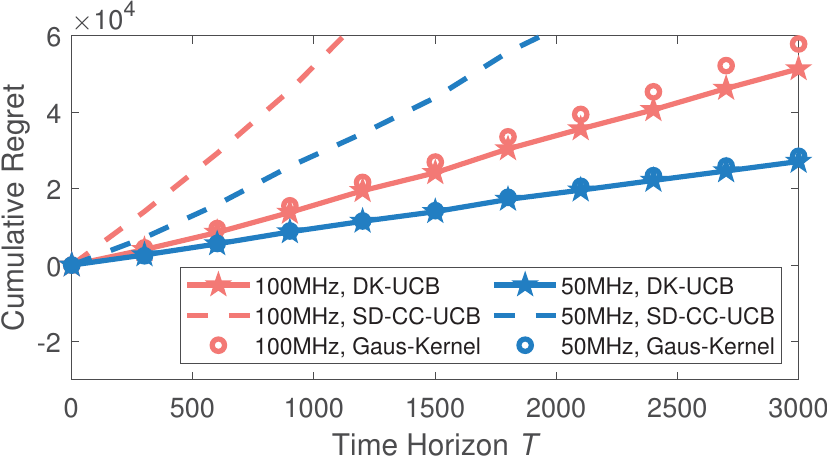}
	\caption{Cumulative regret.}
	\label{Fig_1}
\end{figure}

Our proposed algorithms are compared to other algorithms, i.e., the worst connection swapping (WCS) algorithm \cite{8677293} and the semi-distributed contextual correlated upper confidence bound (SD-CC-UCB) algorithm \cite{CC-UCB}.
WCS is a centralized offline algorithm and requires the instantaneous channel state information (CSI) between all vehicles and BSs, which achieves a near-optimal solution for the NP-hard user association problem.
Without the instantaneous CSI between all vehicles and BSs, SD-CC-UCB is an online algorithm based on the well-known CMAB framework.
SD-CC-UCB partitions the context space evenly and leverages the correlation between contexts for fast convergence.

\begin{figure}[h]
	\centering\includegraphics[scale=0.55]{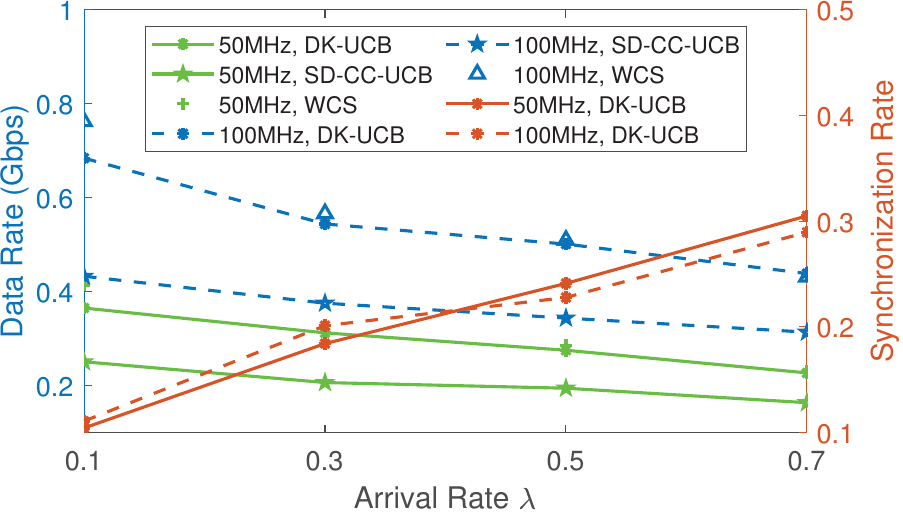}
	\caption{Average transmission rate per vehicle and synchronization rate against vehicle arrival rate.}
	\label{Fig_3}
\end{figure}

Given the vehicle arrival rate of 0.3 (vehicles per second) and a transmit power of 30dBm, Fig.~\ref{Fig_1} shows the cumulative regret, which represents the difference between the best strategy and algorithms.
At the period of 2000, the cumulative regret of the SD-CC-UCB algorithm is approximately 2.3 times higher than that of the DK-UCB algorithm for both 50 MHz and 100 MHz bandwidths. 
DK-UCB leverages the similarity between all the past contexts and current context by the kernel function to estimate the current reward, while each past context and reward contributes to only one partitioned hypercube in SD-CC-UCB.
Gaus-Kernel indicates the kernel function in DK-UCB is replaced by a single Gaussian kernel function $\kappa\big(\mathbf x_{t,i}(a),\hat{\mathbf{x}}_{\hat t,j}(\hat a)\big) = \exp\big(-{|\mathbf x_{t,i}(a) - \hat{\mathbf{x}}_{\hat t,j}(\hat a)|}^2/(2\sigma_{Gaus}^2)\big)$, where $\sigma_{Gaus}$ is a parameter. 
The cumulative regret of Gaus-Kernel is approximately 12.6\% and 5.0\% higher than that of DK-UCB using our proposed kernel function after 3000 periods with different bandwidths, respectively. 
The proposed kernel function evaluates the similarity between contexts in path loss, Doppler spread and concurrent transmissions, better capturing the characteristics of mmWave channels.

\begin{figure}[hpb]
	\centering\includegraphics[scale=0.55]{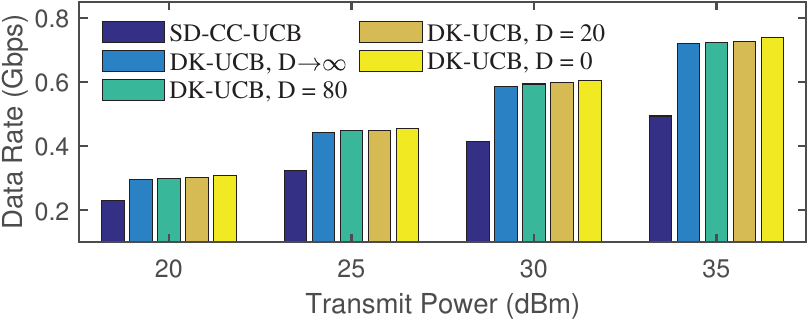}
	\caption{Average transmission rate per vehicle against transmit power.}
	\label{Fig_2}
\end{figure}

With a transmit power of 30dBm, Fig.~\ref{Fig_3} presents the average transmission rate per vehicle and synchronization rate against vehicle arrival rate.
The synchronization rate is defined as the rate of synchronization periods to the whole period.
The transmission rate of DK-UCB is 10.2\% lower than the centralized offline WCS algorithm when the vehicle arrival rate is 0.1.
However, as the vehicle arrival rate increases to 0.7, the situation reverses, and the transmission rate of WCS becomes 1.6\% lower than that of DK-UCB. 
When there are fewer vehicles, fewer contexts and rewards are sampled, resulting in low estimation accuracy.
With a larger number of vehicles, more contexts and rewards become available for estimation, and interference is effectively managed by the kernel function, leading to improved performance.

\begin{figure}[h]
	\centering\includegraphics[scale=0.55]{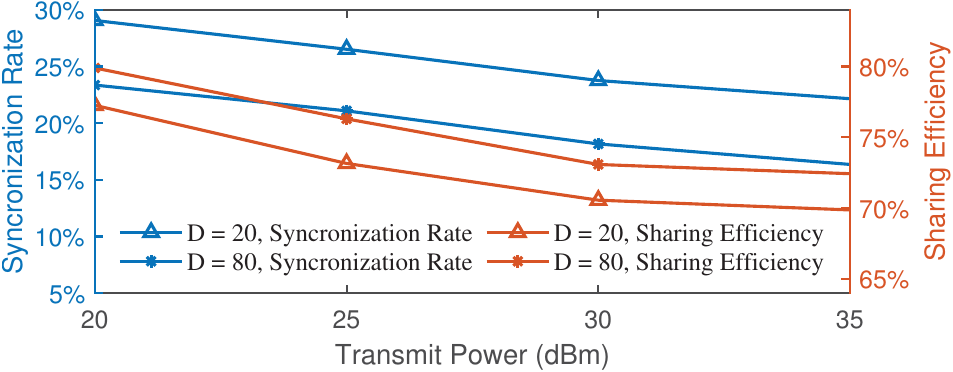}
	\caption{Synchronization rate and sharing efficiency of the DK-UCB algorithm against transmit power and coefficient $D$.}
	\label{Fig_4}
\end{figure}

Fig.~\ref{Fig_2} and Fig.~\ref{Fig_4} present the average transmission rate per vehicle and communication cost given the bandwidth of 100MHz and the vehicle arrival rate of 0.3.
The sharing efficiency is defined as the reduced sharing data rate of DK-UCB within each synchronization by identifying the context subspace in (\ref{subspace_definition}), formally, $1 - |\{\cup_{k\neq i}^{\mathbb{U}(t)}\mathbf S_k^a(t^{k,a}_{syn})\} \cap \mathbb{S}(\mathbf x_{t,i}(a))|\ /\ |\{\cup_{k\neq i}^{\mathbb{U}(t)}\mathbf S_k^a(t^{k,a}_{syn})\}|$.
Although DK-UCB with $D = \infty$ introduces no communication, the average transmission rate is decreased by 2.6\% compared to DK-UCB with $D = 20$ given transmit power of 20dBm. 
Compared to DK-UCB with $D = 0$ triggering synchronizations for all periods and sharing all data within each synchronization, the average transmission rate of DK-UCB with $D = 20$ decreases by 1.0\%-1.6\% across different transmit power levels. 
However, DK-UCB with $D = 20$ introduces only 22.1\%-29.1\% synchronization rate and saves 66.9\%-77.3\% sharing data size within each synchronization across different transmit power levels. 
As shown by the result, the proposed DK-UCB algorithm triggers a synchronization while significant explorations are conducted and shares necessary data, eliminating unnecessary data sharing.
Given the transmit power of 20dBm, the average data rate of DK-UCB with $D = 20$ is 0.9\% higher than that of DK-UCB with $D = 80$, but the synchronization rate is also 5.7\% higher. 
The result indicates that the coefficient $D$ achieves a trade-off between communication cost and estimation accuracy.


%

\section{Conclusion}
In this paper, a DK-UCB algorithm for user association in mmWave vehicular networks is proposed to estimate the transmission rate using past contexts and rewards without extra channel estimations.
We develop kernel methods to capture the nonlinear mapping from a context to the reward and propose a kernel function to capture the propagation characteristics of mmWave signals.
Moreover, DK-UCB incorporates necessary information exchange while vehicles conduct significant explorations, accelerating learning while maintaining reasonable communication costs.

\bibliographystyle{IEEEtran}
\bibliography{reference}

\end{document}